\pdfoutput=1

\documentclass[11pt]{article}

\usepackage[]{emnlp2022}

\usepackage{times}
\usepackage{latexsym}

\usepackage[T1]{fontenc}

\usepackage[utf8]{inputenc}

\usepackage{microtype}

\usepackage{graphicx}
\usepackage{amsmath}
\usepackage{commath}
\usepackage{amsfonts}
\usepackage{multirow}
\usepackage{makecell}
\usepackage{float}

\DeclareUnicodeCharacter{00A0}{ }

\title{CDGP: Automatic Cloze Distractor Generation based on Pre-trained Language Model}

\author{Shang-Hsuan Chiang \and Ssu-Cheng Wang \and Yao-Chung Fan \\
        Department of Computer Science and Engineering, \\National Chung Hsing University, Taichung, Taiwan}

\begin{document}

\maketitle

\begin{abstract}
Manually designing cloze test consumes enormous time and efforts. The major challenge lies in wrong option (distractor) selection. Having carefully-design distractors improves the effectiveness of learner ability assessment. As a result, the idea of automatically generating cloze distractor is motivated. In this paper, we investigate cloze distractor generation by exploring the employment of pre-trained language models (PLMs) as an alternative for candidate distractor generation. Experiments show that the PLM-enhanced model brings a substantial performance improvement. Our best performing model advances the state-of-the-art result from 14.94 to 34.17 (NDCG@10 score). Our code and dataset is available at \href{https://github.com/AndyChiangSH/CDGP}{https://github.com/AndyChiangSH/CDGP}. 

\end{abstract}

\section{Introduction}

\begin{figure}[t]
    \includegraphics[width=\columnwidth]{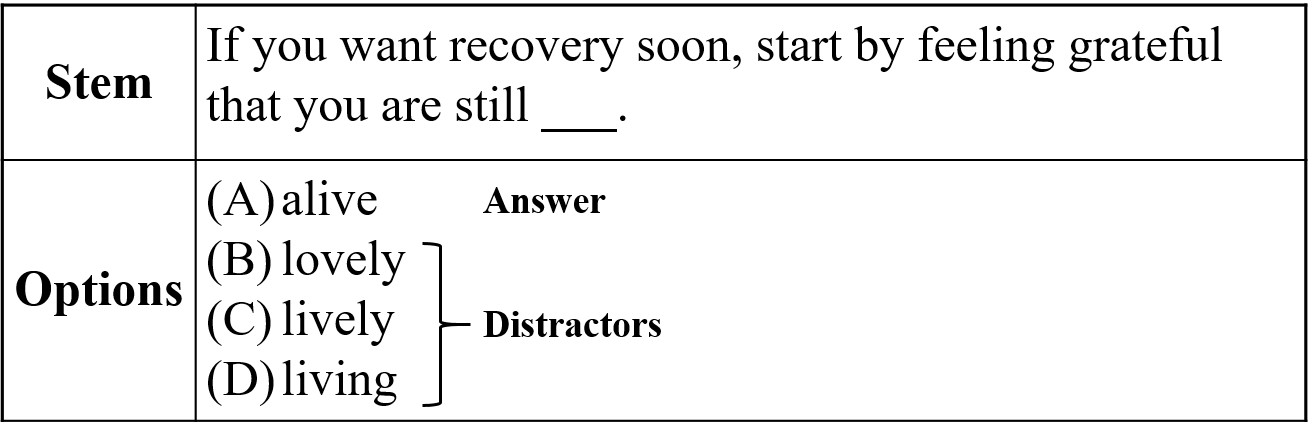}
    \caption{A Cloze Test Example: the challenge to cloze test preparation lies in wrong option selection. A good wrong option selection improve the effectiveness of learner ability assessment.}
    \label{fig:cloze example}
\end{figure}

A cloze test is an assessment consisting of a portion of language with certain words removed (cloze text), where the participant is asked to select the missing language item from a given set of options. Specifically, a cloze question (as illustrated in Figure \ref{fig:cloze example}) is composed by a sentence with a word removed (a blank space) and list of options (one answer and three wrong options).

The cloze test with carefully-design distractors can improve the effectiveness of learner ability assessment. However, manually designing cloze test consumes enormous time and efforts. The major challenge lies in wrong option (distractor) selection. As a result, automatic cloze distractor generation is proposed \cite{Ren_Q.Zhu_2021,kumar2015revup,narendra2013automatic}.

In this paper, we extend the candidate-ranking framework reported in \cite{Ren_Q.Zhu_2021} by exploring the employment of PLMs as an alternative for candidate distractor generation. In this paper, we propose a cloze distractor generation framework called CDGP (Automatic Cloze Distractor Generation based on PLMs) which incorporates a serial of training and ranking strategies to boost the performance of distractor generation based on PLMs.

The contribution of this work is as follows. 

\begin{itemize}
    \item We show that PLM-based methods brings significant performance improvement over the knowledge-driven methods \citep{Ren_Q.Zhu_2021} (generating candidates from Probase \citep{wu2012probase} or Wordnet \citep{miller1995wordnet})
    
    
    \item We conduct evaluation using two benchmarking datasets. The experiment results indicates that our CDGP significantly outperforms the state-of-the-art result \citep{Ren_Q.Zhu_2021}. We advance NDCG@10 score from 19.31 to 34.17 (improving up to 177\%).
    
\end{itemize}
\section{Related Work}

The methods on cloze distractor generation can be sorted into the following two categories. The first category \citep{correia2010automatic,lee2007automatic} is to prepare cloze distractors based on linguistic heuristic rules. The problem with these methods is that the results are far from practically satisfactory. The second category \citep{kumar2015revup,narendra2013automatic} is to construct candidate distractors from domain-specific vocabulary or taxonomies and employ classifiers for selecting final distractors. The results by the methods of this category are still less than satisfactory due to the domain generalization and the generation quality. To improve the quality, \citep{Ren_Q.Zhu_2021} proposes to use knowledge bases (Wordnet \citep{miller1995wordnet} and Probase \citep{wu2012probase}) to analyze the word semantic and hypernym-hyponym relations for generating candidate distractors. In this paper, we explore the employment of 
PLMs as a alternative for the knowledge bases in \citep{Ren_Q.Zhu_2021} and also explore various linguistic features for candidate selection.





\section{Methodology}
\subsection{CDGP Framework}



\begin{figure}[t]
    \centering
    \includegraphics[width=\columnwidth]{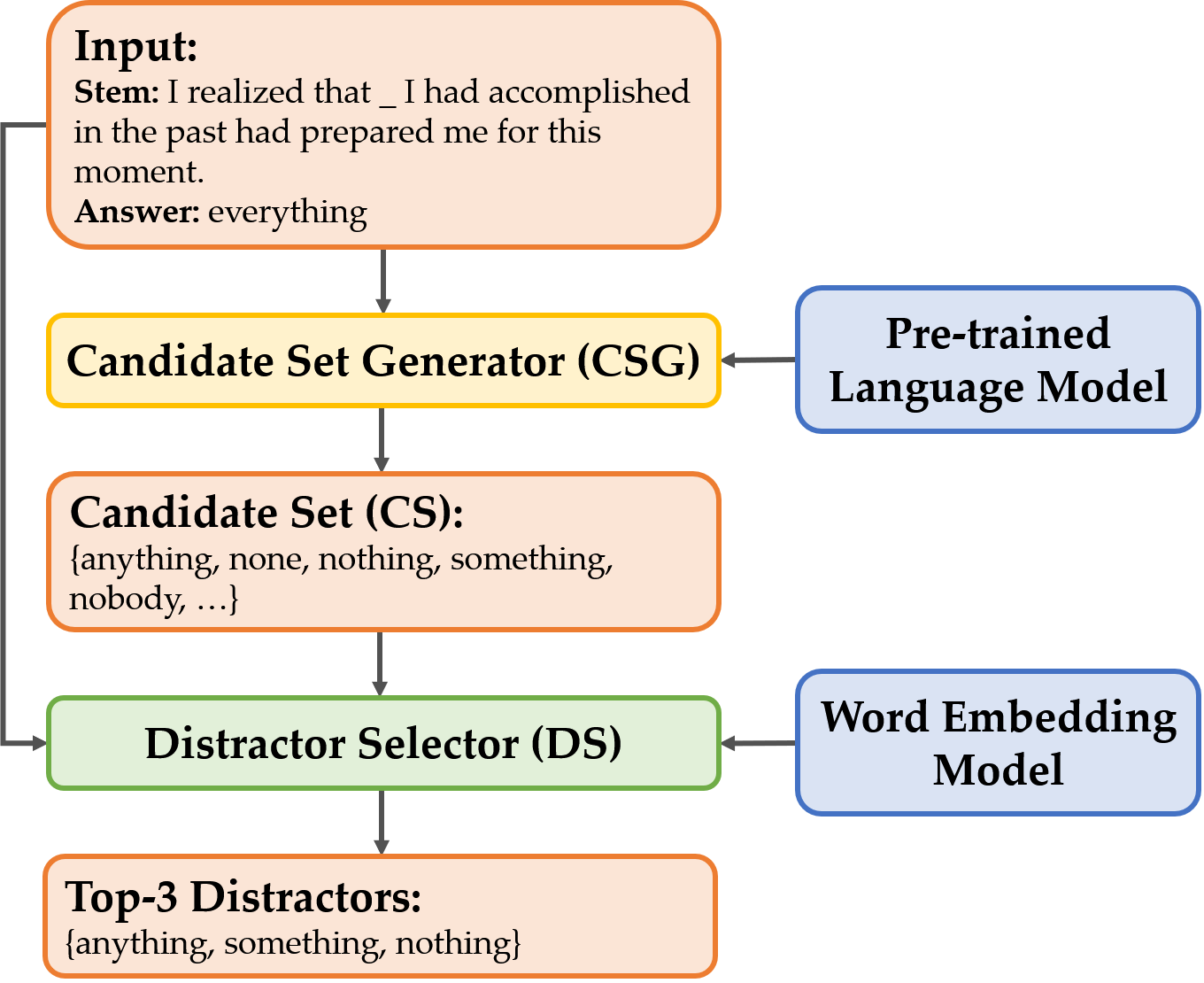}
    \caption{CDGP Framework}
    \label{fig:framework}
\end{figure}

We extend the framework proposed by \citep{Ren_Q.Zhu_2021} by exploring the employment of pre-trained language models as an alternative for candidate distractor generation. Specifically, as illustrated in Figure \ref{fig:framework}, the framework consists of two stages: (1) Candidate Set Generator (CSG) and (2) Distractor Selector (DS). In this paper, we revisit the framework by considering (1) PLMs at CSG and (2) various features at DS.


\subsection{Candidate Set Generator (CSG)}
The input to CSG is a question stem and the corresponding answer. The output is a distractors candidate set of size $k$.

In this study, we use PLM to generate candidates. Let $\mathbb{M}()$ be PLM model. For a given training instance $(S, A, D)$, where $S$ is a cloze stem, $A$ is the answer, and $D$ is a distractor. We explore the following two training setting for generating distractor candidates.






\begin{enumerate}
    \item \textbf{Naive Fine-Tune}: 
    \[\mathbb{M}(S_{\otimes [\mathtt{Mask}]})\rightarrow D\]
    The input is a given stem $S$ with the cloze blank filled in $[\mathtt{Mask}]$ (denoted by $S_{\otimes [\mathtt{Mask}]}$). The idea is to fine-tune the PLMs to predict $D$. The training objective is to find a parameter set $\theta$ minimizing the following loss function 
    \[-log(p(D|S;\theta))\]
    


    \item \textbf{Answer-Relating Fine-Tune}: The input is further concatenated with cloze answer $A$. The idea is to guide the model to refer $A$ to generate $D$. Specifically, 
    \[\mathbb{M}(S_{\otimes [\mathtt{Mask}]}\mathtt{[Sep]}A)\rightarrow D\]
    
    The training objective is to find a parameter set $\theta$ minimizing the following loss function 
    \[-log(p(D|S,A;\theta))\]
    
    
\end{enumerate}

\subsection{Distractor Selector (DS)}
The input to DS is a question stem $S$, an answer $A$, and a candidate set \{$D_i$\} from CSG. We investigate the following features for ranking candidates.


\begin{itemize}
    \item Confidence Score $s_0$: the confidence score of $D_i$ given by the PLM at CSG. Specifically, \[s_0=p(D_i|S,A;\theta)\]
    \item Word Embedding Similarity $s_1$: the word embedding score between $A$ and $D$ given by the cosine similarity between $\vec{A}$ and $\vec{D}$. Specifically, 
    \[ s_1=1-cos(\vec{A},\vec{D_i})\] 
    \item Contextual-Sentence Embedding Similarity $s_2$: the sentence-level cosine similarity between the stem with the blank filled in $A$ (denoted by $\vec{S}_{\otimes A}$) and the stem with the blank filled in $D$ (denoted by $\vec{S}_{\otimes D_i}$).
    \[ s_2=1-cos(\vec{S}_{\otimes A},\vec{S}_{\otimes D_i})\]     
    
    \item POS match score $s_3$: the POS (part-of-speech) matching indicator. $s_3=1$, if $A$ and $D_i$ has the same POS tag. Otherwise $s_3=0$. 
\end{itemize}

The final score of a distractor $D_i$ is then computed by a weighted sum over the individual score with MinMax normalization.
\[
 score(D_i)=\sum_{i=0}^3 w_i\cdot \texttt{MinMax-Norm}(s_i)
\]

Distractors with Top-3 scores are selected as the final resultant distractors.

\section{Performance Evaluation}

\subsection{Dataset}
To validate the performance of our methodology, we use the following two datasets.

\begin{table}[t]
\centering
\resizebox{\linewidth}{!}{
\begin{tabular}{|l|ccc|ccc|ccc|}
\hline
\multirow{2}{*}{Dataset} & \multicolumn{3}{c|}{CLOTH-M}                                  & \multicolumn{3}{c|}{CLOTH-H}                                  & \multicolumn{3}{c|}{CLOTH (Total)}                              \\ \cline{2-10} 
                         & \multicolumn{1}{c|}{train} & \multicolumn{1}{c|}{dev}  & test & \multicolumn{1}{c|}{train} & \multicolumn{1}{c|}{dev}  & test & \multicolumn{1}{c|}{train} & \multicolumn{1}{c|}{dev}   & test  \\ \hline
\#passages               & \multicolumn{1}{c|}{2341}  & \multicolumn{1}{c|}{355}  & 355  & \multicolumn{1}{c|}{3172}  & \multicolumn{1}{c|}{450}  & 478  & \multicolumn{1}{c|}{5513}  & \multicolumn{1}{c|}{805}   & 813   \\ \hline
\#questions              & \multicolumn{1}{c|}{22056} & \multicolumn{1}{c|}{3273} & 3198 & \multicolumn{1}{c|}{54794} & \multicolumn{1}{c|}{7794} & 8318 & \multicolumn{1}{c|}{76850} & \multicolumn{1}{c|}{11067} & 11516 \\ \hline
Vocab. size              & \multicolumn{3}{c|}{15096}                                    & \multicolumn{3}{c|}{32212}                                    & \multicolumn{3}{c|}{37235}                                      \\ \hline
Avg. \#sentence          & \multicolumn{3}{c|}{16.26}                                    & \multicolumn{3}{c|}{18.92}                                    & \multicolumn{3}{c|}{17.79}                                      \\ \hline
Avg. \#words             & \multicolumn{3}{c|}{242.88}                                   & \multicolumn{3}{c|}{365.1}                                    & \multicolumn{3}{c|}{313.16}                                     \\ \hline
\end{tabular}}
\caption{The statistics of the training, developing and testing datasets of CLOTH-M (middle school),\\CLOTH-H (high school). \citep{xie2017large}}
\label{table:cloze dataset}
\end{table}

\begin{table}[t]
\centering
\resizebox{\linewidth}{!}{
\begin{tabular}{|l|cc|cc|c|}
\hline
\multirow{2}{*}{Dataset} & \multicolumn{2}{c|}{Short-term}    & \multicolumn{2}{c|}{Long-term}     & \multicolumn{1}{l|}{} \\ \cline{2-6} 
                         & \multicolumn{1}{c|}{GM}    & STR   & \multicolumn{1}{c|}{MP}    & LTR   & O                     \\ \hline
CLOTH                    & \multicolumn{1}{c|}{0.265} & 0.503 & \multicolumn{1}{c|}{0.044} & 0.180 & 0.007                 \\ \hline
CLOTH-M                  & \multicolumn{1}{c|}{0.330} & 0.413 & \multicolumn{1}{c|}{0.068} & 0.174 & 0.014                 \\ \hline
CLOTH-H                  & \multicolumn{1}{c|}{0.240} & 0.539 & \multicolumn{1}{c|}{0.035} & 0.183 & 0.004                 \\ \hline
\end{tabular}}
\caption{The question type statistics of 3000 sampled questions where GM, STR, MP, LTR and O denotes grammar, short-term-reasoning, matching paraphrasing, long-term-reasoning and others respectively. \citep{xie2017large}}
\label{table:cloze dataset details}
\end{table}

\begin{table}[t]
\centering
\setlength{\tabcolsep}{7mm}{\small
\begin{tabular}{|l|c|}
\hline
Data Split & \# of questions \\ \hline
total   & 2880        \\ \hline
train   & 2321        \\ \hline
valid   & 300         \\ \hline
test    & 259         \\ \hline
\end{tabular}}
\caption{DGen statistics. \citep{Ren_Q.Zhu_2021}}
\label{table:dgen question numbers}
\end{table}

\begin{table}[htb]
\centering\resizebox{\linewidth}{!}{
\setlength{\tabcolsep}{1mm}{
\begin{tabular}{|l|c|c|c|c|c|}
\hline
Domain        & Total & Science & Vocab. & \begin{tabular}[c]{@{}c@{}}Commen\\Sense\end{tabular} & Trivia \\ \hline
\# of questions   & 2880  & 758     & 956    & 706          & 460    \\ \hline
\#of distractors & 3.13  & 3.00    & 3.99   & 3.48         & 2.99   \\ \hline
\end{tabular}}}
\caption{DGen statistics in different domains. \citep{Ren_Q.Zhu_2021}}
\label{table:dgen question areas}
\end{table}

\begin{itemize}
    \item \textbf{CLOTH datatset} \citep{xie2017large} The dataset comes from English cloze exercises. The datasets consists of a passage with cloze stems, answers and distractors. The data statistics are summarized in Table \ref{table:cloze dataset} and Table \ref{table:cloze dataset details}.
    
    \item \textbf{DGen dataset} \cite{Ren_Q.Zhu_2021} The DGen dataset released by \citep{Ren_Q.Zhu_2021}, which is a reorganized dataset from SciQ \citep{welbl2017crowdsourcing} and MCQL \citep{liang2018distractor}. We compare our methods with the SOTA method \citep{Ren_Q.Zhu_2021} based on this dataset. The data statistics are listed in Table \ref{table:dgen question numbers} and Table \ref{table:dgen question areas}.
\end{itemize}



\subsection{Implementation Details}
We select bert-base-uncased \citep{DBLP:journals/corr/abs-1810-04805} as the default PLM. We use Adam optimizer with an initial learning rate setting to 0.0001. We set the PLM maximal input length to 64. The default batch size is set to 64. All models are trained with NVIDIA® Tesla T4.


For computing word embedding similarity in DS, we use the fasttext model \citep{bojanowski2016enriching} as the default embedding model. The fasttext is trained with the cbow setting. The minimal and maximal n-gram parameter are set to 3 and 6. The vector dimension is set to 100. The initial learning rate is 0.05. In addition, the size of distractor candidate set $k$ is set to 10 as a default value.



\subsection{Evaluation Metric}
\paragraph{Automatic Score}
We use the same setting of \citep{Ren_Q.Zhu_2021}; the models are compared by the following automatic scores: Precision (P@1), F1 score (F1@3, F1@10), Mean Reciprocal Rank (MRR@10), and Normalized Discounted Cumulative Gain (NDCG@10).


\subsection{Evaluation Results}
\subsubsection{Results on DGen}

\begin{table*}[t]
\centering
\begin{tabular}{lcccc}
\hline
\multicolumn{1}{c}{\color[HTML]{222222} \textbf{Models}} & {\color[HTML]{222222} \textbf{P@1}} & {\color[HTML]{222222} \textbf{F1@3}} & {\color[HTML]{222222} \textbf{MRR@10}} & {\color[HTML]{222222} \textbf{NDCG@10}} \\ \hline
\textbf{DGen (Wordnet CSG)} & 9.31 & 7.71 & 14.34 & 14.94 \\
\textbf{DGen (Probase CSG)} & 10.85 & 9.19 & 17.51 & 19.31 \\
\textbf{DGen (w/o CSG)} & 5.01 & 5.59 & 9.28 & 11.6  \\ \hline
\textbf{CDGP (BERT)} & 10.81 & 7.72 & 18.15 & 24.47 \\
\textbf{CDGP (SciBERT)} & \textbf{13.13} & \textbf{12.23} & \textbf{25.12} & \textbf{34.17} \\
\textbf{CDGP (RoBERTa)} & {13.13} & {9.65} & {19.34} & {24.52} \\ 
\textbf{CDGP (BART)} & 8.49 & 8.24 & 16.01 & 22.66 \\
\hline
\end{tabular}
\caption{Comparison Results: Comparing CDGP with the DGen \cite{Ren_Q.Zhu_2021}}
\label{table:compare with dgen}
\end{table*}
In this set of experiment, our goal is to compare our method with the SOTA method \cite{Ren_Q.Zhu_2021}. Table \ref{table:compare with dgen} shows the comparison results. In addition to the BERT model, we also report our CDGP variants based on (SciBERT, RoBERTa, and BART). From Table \ref{table:compare with dgen}, it can be seen that the NDCG@10 of CDGP with SciBERT was improved from 19.31 to 34.17, surpassing the existing SOTA method by 77\%. 

An interesting finding here is that in this set of experiment, we see CDGP using SciBERT show the best results. We think this confirms the domain matchesness between DGen dataset. Note SciBERT which is pre-trained based on science literature and DGen is a dataset related to scientific domains. 

\subsubsection{Results on CLOTH dataset}
In this experiment, we evaluate the performance of our models on CLOTH dataset and conduct ablation studies for our CDGP model.

\paragraph{Comparing Fine-Tuning Strategy}
In this set of experiment, we compare the performance of naive fine-tuning and answer-relating fine-tuning. The results are presented in Table \ref{table:with ans or not}.

\begin{table}[t]
\centering\resizebox{\linewidth}{!}{
\begin{tabular}{lccccc}
\hline
{\color[HTML]{222222} \textbf{Models}}                                         & {\color[HTML]{222222} \textbf{P@1}} & {\color[HTML]{222222} \textbf{F1@3}} & {\color[HTML]{222222} \textbf{F1@10}} & {\color[HTML]{222222} \textbf{MRR@10}}   & {\color[HTML]{222222} \textbf{\begin{tabular}[c]{@{}c@{}}NDCG\\ @10\end{tabular}}} \\
\hline
{\color[HTML]{222222} \textbf{Naive}}                                                    & {\color[HTML]{222222} 12.60} & {\color[HTML]{222222} 10.00} & {\color[HTML]{222222} 12.45}          & {\color[HTML]{222222} 22.70}          & {\color[HTML]{222222} 30.32}                                                       \\
{\color[HTML]{222222} \textbf{\begin{tabular}[c]{@{}l@{}}Answer\\ Relating\end{tabular}}} & {\color[HTML]{222222} \textbf{18.50}} & {\color[HTML]{222222} \textbf{13.80}} & {\color[HTML]{222222} \textbf{15.37}} & {\color[HTML]{222222} \textbf{29.96}} & {\color[HTML]{222222} \textbf{37.82}}\\ \hline                                          
\end{tabular}}
\caption{The Results of Naive and Answer-Relating Fine-Tuning Comparison}
\label{table:with ans or not}
\end{table}


From the above results, it can be observed that the overall score of answer-relating fine-tuning is higher than that of naive fine-tuning. Therefore, we select answer-relating fine-tuning as a default fine-tuning strategy.

\paragraph{Comparing Pre-trained Language Models}

In this set of experiment, we experiment with using different pre-trained language models.  The following are the pre-trained language models used in the experiments.
(1) BERT \citep{DBLP:journals/corr/abs-1810-04805}, (2) SciBERT \citep{beltagy-etal-2019-scibert}, (3) RoBERTa \citep{DBLP:journals/corr/abs-1907-11692}, (4) BART \citep{DBLP:journals/corr/abs-1910-13461}.

\begin{table}[t]
\centering\resizebox{\linewidth}{!}{
\begin{tabular}{lccccc}
\hline
{\color[HTML]{222222} \textbf{Models}} & {\color[HTML]{222222} \textbf{P@1}} & {\color[HTML]{222222} \textbf{F1@3}} & {\color[HTML]{222222} \textbf{F1@10}} & {\color[HTML]{222222} \textbf{MRR@10}} & {\color[HTML]{222222} \textbf{\begin{tabular}[c]{@{}c@{}}NDCG\\ @10\end{tabular}}} \\ \hline
\textbf{\begin{tabular}[c]{@{}l@{}}BERT\end{tabular}} & \textbf{18.50} & \textbf{13.80} & \textbf{15.37} & \textbf{29.96} & \textbf{37.82} \\
\textbf{\begin{tabular}[c]{@{}l@{}}SciBERT\end{tabular}} & 8.10 & 9.13 & 12.22 & 19.53 & 28.76 \\
\textbf{\begin{tabular}[c]{@{}l@{}}RoBERTa\end{tabular}} & 10.50 & 9.83 & 10.25 & 20.42 & 28.17 \\
\textbf{\begin{tabular}[c]{@{}l@{}}BART\end{tabular}} & 14.20 & 11.07 & 11.37 & 24.29 & 31.74 \\
\hline     
\end{tabular}}
\caption{Results on Comparing the Employment of Different Pre-trained Language Models (fine-tuned with CLOTH dataset)}
\label{table:diff model}
\end{table}

Table \ref{table:diff model} shows the comparison result. Through this experiment, we see that the BERT model has the most outstanding performance, so we use the BERT model for subsequent experiments.

\paragraph{Comparing DS Factors}


There are four scoring factors in DS, namely $s_0$ (confidence score), $s_1$ (word embedding similarity), $s_2$ (contextual sentence similarity) and $s_3$ (part-of-speech match score). In this experiment, we adjust the weighting of each scoring index of DS (from $w_0$ to $w_3$), and compare the difference of using different weight ratios. Table \ref{table:ds parameter} shows the experiment results.

\begin{table}[t]
\centering\resizebox{\linewidth}{!}{
\begin{tabular}{llll|cccc}
\hline
{\color[HTML]{222222} \textbf{\begin{tabular}[c]{@{}l@{}} $w_0$ \end{tabular}}} & {\color[HTML]{222222} \textbf{\begin{tabular}[c]{@{}l@{}}$w_1$\end{tabular}}} & {\color[HTML]{222222} \textbf{\begin{tabular}[c]{@{}l@{}}$w_2$\end{tabular}}} & {\color[HTML]{222222} \textbf{\begin{tabular}[c]{@{}l@{}}$w_3$\end{tabular}}} & {\color[HTML]{222222} \textbf{P@1}} & {\color[HTML]{222222} \textbf{F1@3}} & {\color[HTML]{222222} \textbf{MRR@10}} & {\color[HTML]{222222} \textbf{\begin{tabular}[c]{@{}c@{}}NDCG\\ @10\end{tabular}}} \\ \hline
\textbf{0.25} & \textbf{0.25} & \textbf{0.25} & \textbf{0.25} & 18.50 & 13.80 & 29.96 & 37.82 \\
\textbf{0.4} & \textbf{0.2} & \textbf{0.2} & \textbf{0.2} & \textbf{19.40} & 15.33 & 31.11 & 39.12 \\
\textbf{0.6} & \textbf{0.15} & \textbf{0.15} & \textbf{0.1} & 19.30 & \textbf{15.50} & \textbf{31.26} & 39.49\\
\textbf{0.8} & \textbf{0.05} & \textbf{0.05} & \textbf{0.1} & 18.90 & 15.43 & 30.88 & \textbf{39.56} 
\\
\hline
\end{tabular}}
\caption{Distractor Selector Features Weighting Comparison}
\label{table:ds parameter}
\end{table}


From the results in Table \ref{table:ds parameter}, we see that if the weights of $s_1$ and $s_2$ is adjusted lower, a better distractor generation performance is observed, but if they are set too low, the performance starts to degrade. 


After the experiments, we see that the DS weights setting to (0.6, 0.15, 0.15, 0.1) show the best performance. We use this weighting setting as default values for other experiments.


\paragraph{Comparing w/o CDGP Components}

Through the above experiment studies, we obtain the besting parameter settings for CDGP. In order to prove the effectiveness of the CDGP design, in this set of experiments, we compare the use or not of each component in the framework. Table \ref{table:with frame or not} presents the experimental results.

\begin{table}[t]
\centering\resizebox{\linewidth}{!}{
\begin{tabular}{lccccc}
\hline
{\color[HTML]{222222} \textbf{Methods}} & {\color[HTML]{222222} \textbf{P@1}} & {\color[HTML]{222222} \textbf{F1@3}} & {\color[HTML]{222222} \textbf{F1@10}} & {\color[HTML]{222222} \textbf{MRR@10}} & {\color[HTML]{222222} \textbf{\begin{tabular}[c]{@{}c@{}}NDCG\\ @10\end{tabular}}} \\ \hline
\textbf{CSG+DS} & \textbf{19.30} & \textbf{15.50} & \textbf{15.37} & \textbf{31.26} & \textbf{39.49}  \\
\textbf{CSG} & 18.50 & 14.90 & 15.37 & 30.57& 38.73  \\
\textbf{DS} & 4.00 & 6.43 & 5.05 & 12.02 & 19.12  \\
\textbf{None} & 4.10 & 6.03 & 5.05 & 11.81 & 18.65  \\ \hline
\end{tabular}}
\caption{Ablation study on CDGP components}
\label{table:with frame or not}
\end{table}


From the results, we can see that the whole CDGP framework (CSG+DS with $(w_0, w_1, w_2, w_3)=(0.6, 0.15, 0.15, 0.1)$)  shows the best performing results compared with the options using only one or none of the components. Furthermore, we see that using only CSG improves the performance (107.7\%, in terms of NDCG@10, compared with \textit{none} scheme (which uses BERT's MLM capability to have distractor candicate without any fine-tuning), while using only DS brings slightly performance improvement (2.5\%). Such results indicate that the major performance improvement comes from the CSG employment.

\subsubsection{Result on Human Evaluation}

We also recruit 40 human evaluators from our campus. The evaluation process is as follows. First, the evaluator takes a cloze exam (a passage with 10 cloze multiple choice questions). The passages are randomly selected from the CLOTH dataset. For a selected passage, we keep five original questions and replace the rest five questions with the generation results by our model. Our goal is to observe the answering correct rate over the manually designed distractors and the automatically designed distractors. Furthermore, we also ask the evaluators to exam the quality of the generated distractors. Specifically, after the exam, we ask (1) the evaluators to guess which questions are generated by CDGP and (2) rank the distractor difficulty by Likert scale ranging from 1-5.

\paragraph{Answering Correct Rate} We find that the correct rate of the human cloze questions is 50.5\%, while the correct rate of CDGP questions is 66\%. The correct rate of CDGP questions is slightly higher than that of human questions, which shows that the difficulty of CDGP distractors is slight easier than that of human questions. Improving and controlling the difficulty of automatically generated distractors will be an interesting future work direction.

\paragraph{Distinguishing Human-design or CDGP Question}
In the test of judging whether a question is a CDGP question, the correct rate of the evaluators' guess is 53\%, which nearly to a random guess, showing that the evaluator cannot effectively distinguish between human and CDGP questions.

\begin{figure}[t]
    \centering
    \includegraphics[width=.55\columnwidth]{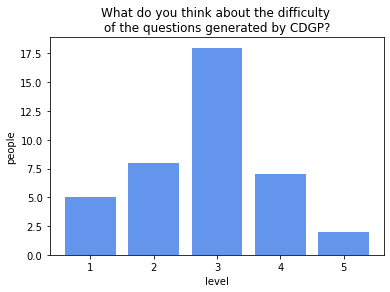}
    \caption{The testers' feedback on the difficulty of the questions generated by CDGP (1: easiest, 5: most difficult)}
    \label{fig:human2}
\end{figure}

\paragraph{Examining Difficulty of Generated Distractors}
From the tester feedback, as shown in Figure \ref{fig:human2}, the testers' ratings of difficulty are normally distribution, indicating that the difficulty level of the questions is moderate. It can be seen that the performance of CDGP questions is close to that of manual-design questions, which confirms that CDGP can assist in the cloze distractor preparation.




\section{Conclusion}\vspace{-2mm}
Our study indicates that PLM-based candidate distractor generator is a better alternative for knowledge-based component. The experiment results show that our model significantly surpassed the SOTA method, demonstrating the effectiveness of PLM-based distractor generation on Cloze Test. Also, the result shows that using domain-specific PLM will further boost the generation quality.



\section{Limitations}\vspace{-1mm}
The major limitation for this study is that the current evaluation on the test dataset cannot truly reflect the distractor generation quality. A mismatch with the ground truth distractors do not imply the generated distractor is not a feasible one. Also, we have no way to control the difficulty and the correctness of distractor generation.
\section*{Acknowledgement}
This work is supported by NSTC 110-2634-F-005-006-project Smart Sustainable New Agriculture Research Center (SMARTer), NSTC Taiwan Project under grant 109-2221-E-005-058-MY3, and Delta Research Center, Delta Electronics, Inc. We thank to National Center for High-performance Computing (NCHC) of National Applied Research Laboratories (NARLabs) in Taiwan for providing computational and storage resources.
\bibliographystyle{acl_natbib}
\bibliography{Reference}

\newpage
\appendix

\end{document}